\documentclass[11pt]{article}
\usepackage[T1]{fontenc}
\usepackage[utf8]{inputenc}
\usepackage{newtxtext,newtxmath}
\usepackage[margin=1in]{geometry}
\usepackage{microtype}
\usepackage{graphicx}
\usepackage{booktabs}

\usepackage{xcolor}
\usepackage[numbers,sort&compress]{natbib}
\usepackage[font=small,labelfont=bf]{caption}
\usepackage{enumitem}
\setlist{nosep,leftmargin=1.5em}
\usepackage{listings}
\usepackage[colorlinks,linkcolor=black,citecolor=black,urlcolor=blue!50!black]{hyperref}
\hypersetup{pdftitle={MemoryTax: The Bitter Lesson of Agent Memory}, pdfauthor={Zedong Peng}}

\newcommand{\figrule}{\color{black!18}\hrule height 0.4pt}
\title{MemoryTax: The Bitter Lesson of Agent Memory}
\author{Zedong Peng\\Shanghai Jiao Tong University\\\texttt{zedongpeng1@gmail.com}}
\date{}

\begin{document}
\maketitle

\begin{center}
\large\href{https://zedongpeng.com/memorytax/}{Interactive version: \textbf{zedongpeng.com/memorytax}}
\end{center}
\vspace{-0.4em}

\begin{abstract}
Agent memory systems such as Mem0 and Zep decide what to keep before any question arrives; coding agents keep the raw history and let the model search it with its own tools. We ask whether designed memory still helps once the model is strong enough to search for itself. On one LoCoMo conversation (419 turns, 152 questions), we run four memory methods (full context, BM25, Mem0, Zep) and three coding agents (Pi, Codex, Claude Code), each with GPT-4o-mini, GPT-6 Luna and Claude Opus 5.5, and record what every rollout costs. This is a short empirical study, with one conversation and one run per cell. Agent memory is a costly patch for a weak model: Mem0 beats full context only with GPT-4o-mini, and its build alone costs more than a whole full-context rollout. Agent memory also has a ceiling, set by the model that writes it: a stronger reader does not recover what a weaker writer left out. With Claude Opus 5.5, Pi tops the cost--accuracy frontier, scoring 99\% for less than half the cost of full context; its one miss is a wrong answer key. Part of that lead comes from photo captions that only the agents could read: on the 88 questions whose evidence has no photo, Pi and full context differ by one answer.
\end{abstract}

\section{Introduction}

Agent memory systems such as Mem0 and Zep decide what to keep before any question arrives: the model extracts facts or builds a graph from the conversation, and each question sees a few retrieved pieces~\citep{chhikara2025mem0,rasmussen2025zep}. Coding agents and recursive language models keep the raw history instead and let the model search it with its own tools~\citep{pi2026,codex2026,claudecode2026,zhang2025rlm}.

The bitter lesson is that knowledge built into a system helps at first, then plateaus, while general methods that scale with computation win~\citep{sutton2019bitter}. Agent memory builds in what to keep. Does it still help once the model is strong enough to search for itself?

\begin{figure}[t]
\centering
\figrule\par\vspace{2pt}
\includegraphics[width=0.68\linewidth]{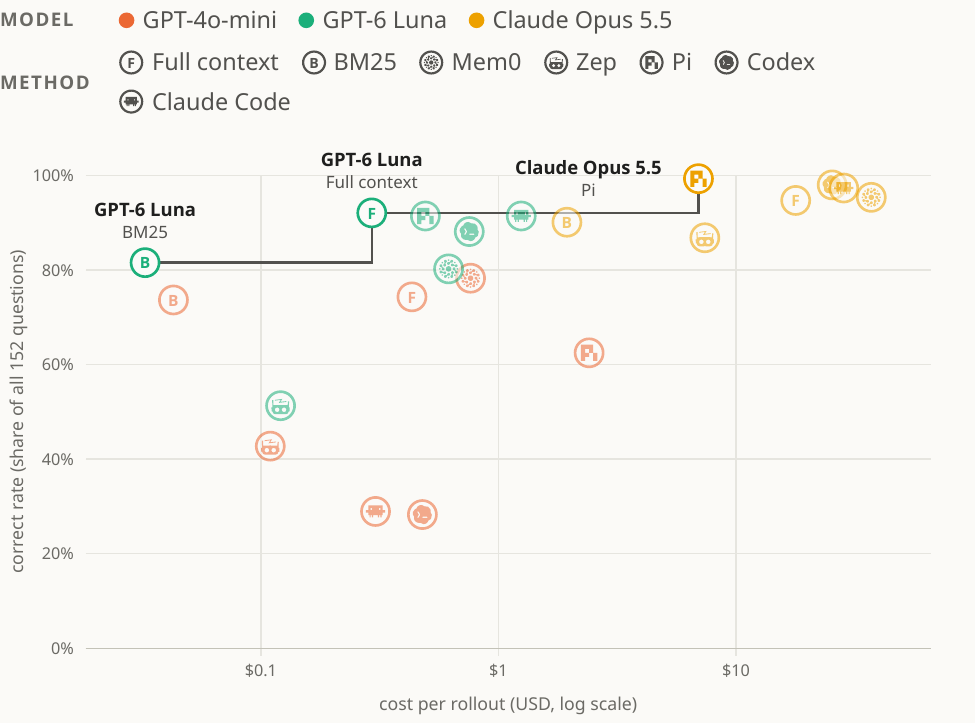}\par\vspace{2pt}\figrule
\caption{Cost and correct rate per rollout. Each badge is one cell's whole run: the build, then all 152 questions. Colour is the model, the badge the method's mark, and a letter in the badge marks a method without one (F full context, B BM25). The stepped line is the highest correct rate reached at or below each cost, and the labelled cells lie on it. Cost is on a log axis.}
\label{fig:frontier}
\end{figure}

We run four memory methods (full context, BM25~\citep{robertson2009bm25}, Mem0 and Zep) and three coding agents (Pi, Codex and Claude Code), each with GPT-4o-mini, GPT-6 Luna and Claude Opus 5.5, on a conversation from LoCoMo~\citep{maharana2024locomo}, and measure what each run costs. We find:

\begin{itemize}
\item \textbf{Agent memory is a costly patch for a weak model.} Mem0 beats full context only with GPT-4o-mini, and its build alone costs more than full context's whole run (Section~\ref{sec:patch}).
\item \textbf{Agent memory has a ceiling, set by the model that writes it.} A stronger reader does not recover what a weaker writer left out (Section~\ref{sec:ceiling}).
\item \textbf{With Claude Opus 5.5, Pi tops the cost--accuracy frontier} (Figure~\ref{fig:frontier}). It answers every question whose answer key is sound, for less than half what full context costs. Where the two read the same text they differ by one answer; the rest of Pi's lead comes from photo captions that only the agents could read (Section~\ref{sec:pi}).
\end{itemize}

The study is small: one conversation, one run per cell, and a GPT-4o-mini judge. Section~\ref{sec:discussion} lists what it cannot show. Within those limits, the results suggest that a strong model needs less designed memory and better tools for searching the raw record.

\section{Setup}
\label{sec:setup}

We run 21 cells, seven methods with three models each, on LoCoMo conv-26~\citep{maharana2024locomo}: 419 turns in 19 sessions and 152 single-hop, multi-hop, temporal and open-domain questions. LoCoMo's adversarial questions are left out, as in Mem0's evaluation~\citep{chhikara2025mem0}. The history fits every model's context window. Among the ten LoCoMo conversations, conv-26's per-method accuracies track those of the full set closely. Table~\ref{tab:methods} lists the methods.

\begin{table}[t]
\centering
\caption{Methods: what each stores and what the model sees when it answers.}
\label{tab:methods}
\small
\begin{tabular}{@{}lll@{}}
\toprule
\textbf{Method} & \textbf{Stored} & \textbf{Model sees} \\
\midrule
Full context & raw history & all of it \\
BM25 & raw turns & top-8 turns, $\pm$2 \\
Mem0 & extracted facts & top-200 facts \\
Zep & temporal graph & top-20 edges \\
Pi, Codex, Claude Code & raw history, one JSON file & what its tools read \\
\bottomrule
\end{tabular}
\end{table}

The memory methods answer one question at a time with Mem0's 2025 answer template; Mem0 is OSS 2.1.0 and Zep is Graphiti OSS 0.30.2, both with default extraction and retrieval~\citep{chhikara2025mem0,rasmussen2025zep}. The agents, Pi 1.1.0, Codex 0.159.2 and Claude Code 2.1.295~\citep{pi2026,codex2026,claudecode2026}, start a fresh session in a fresh container for each question, with the conversation as one file, their default tools, and the template's answer rules in the prompt (Appendix~\ref{app:prompts}).

The models go from weak to strongest: GPT-4o-mini, GPT-6 Luna, Claude Opus 5.5. The last two run at reasoning effort high, the effort coding agents are usually run at~\citep{arena2026harnesstax}; GPT-4o-mini has no effort setting. In each cell one model writes the store (Mem0, Zep) and answers every question, or drives the agent.

\paragraph{Photos.} Some LoCoMo turns share a photo, and the official file holds a caption of it next to the turn text. The agents get the official file, captions included; the memory methods were given the turn text only. 116 of the 419 turns share a photo, and 64 of the 152 questions have evidence in such a turn. Section~\ref{sec:pi} measures what this changes.

\paragraph{Measurement.} A rollout is one run over the conversation: the build, then the 152 questions. GPT-4o-mini grades every answer with Mem0's 2025 judge rubric~\citep{chhikara2025mem0}; empty answers count as wrong. Cost is the list price of every model call, LLM and embedding (text-embedding-3-small), with the judge excluded. A turn is one LLM call: one per question for the memory methods, one per step of an agent's loop, and Mem0's and Zep's calls to build their store. Tokens are those calls' input and output tokens, cached input included; embeddings are not counted. Each cell runs once.

\section{Agent memory is a costly patch for a weak model}
\label{sec:patch}

With GPT-4o-mini, Mem0 scores highest. It scores 78\%, against 74\% for both full context and BM25; Zep scores 43\%. The agents score 63\% (Pi), 28\% (Codex) and 29\% (Claude Code): Codex opens the conversation file for three questions in four and Claude Code almost never, and otherwise they answer from the question alone.

With the two stronger models, the patch stops helping. With GPT-6 Luna, full context scores 92\%, BM25 82\%, Mem0 80\% and Zep 51\%; Zep's graph holds 128 facts from the 19 sessions, against 195 with Claude Opus 5.5. With Claude Opus 5.5, Mem0 and full context both score 95\% and Zep 87\%, and the agents score highest: Pi 99\%, Codex 98\% and Claude Code 97\%, a lead that comes mostly from questions about photos (Section~\ref{sec:pi}). Earlier work also finds that plain retrieval keeps up with elaborate memory~\citep{wu2026basics} and that rankings between memory systems change with the model~\citep{wang2026memdelta}.

\begin{figure}[t]
\centering
\figrule\par\vspace{2pt}
\includegraphics[width=0.68\linewidth]{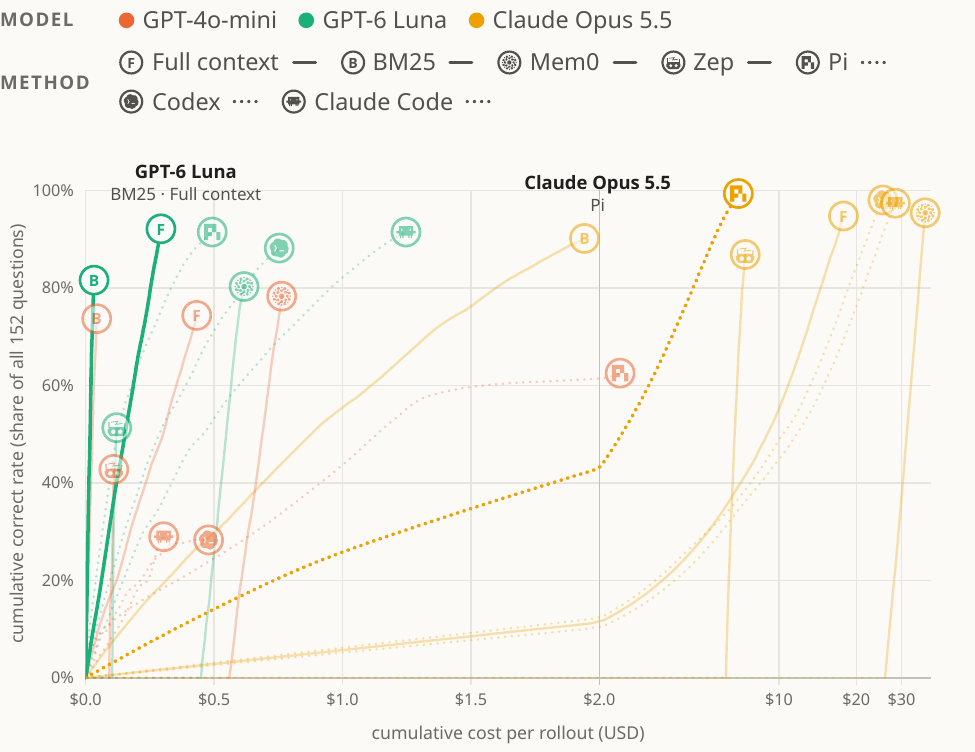}\par\vspace{2pt}\figrule
\caption{Cumulative cost--correct curves. Each curve is one rollout: the build, then the questions from cheapest to most expensive (equal costs grouped), adding up cost and correct answers; correct answers are divided by 152, so a wrong answer adds cost only. Colour is the model, the badge the method, and agents are dotted. The cost axis is linear to \$2 and logarithmic above. Curves use a nine-point moving average, and the badges mark the unsmoothed totals of Table~\ref{tab:results}. Strong lines are cells that no other cell beats on both cost and correct answers.}
\label{fig:cumulative}
\end{figure}

The patch is not free. Mem0's build alone costs more than full context's whole rollout with the same model: \$0.56 against \$0.43 with GPT-4o-mini, \$0.45 against \$0.29 with GPT-6 Luna and \$25.88 against \$17.84 with Claude Opus 5.5 (Figure~\ref{fig:cumulative}). Its questions are cheaper, so the build would pay for itself after 372, 545 and 596 questions on the same history; LoCoMo asks 152. Mem0 and Zep are never on the frontier of Figure~\ref{fig:frontier}, which runs through BM25 and full context with GPT-6 Luna and Pi with Claude Opus 5.5. Paying to build memory that a stronger model does not need is a memory tax.

\section{Agent memory has a ceiling}
\label{sec:ceiling}

\begin{figure}[t]
\centering
\figrule\par\vspace{2pt}
\includegraphics[width=0.68\linewidth]{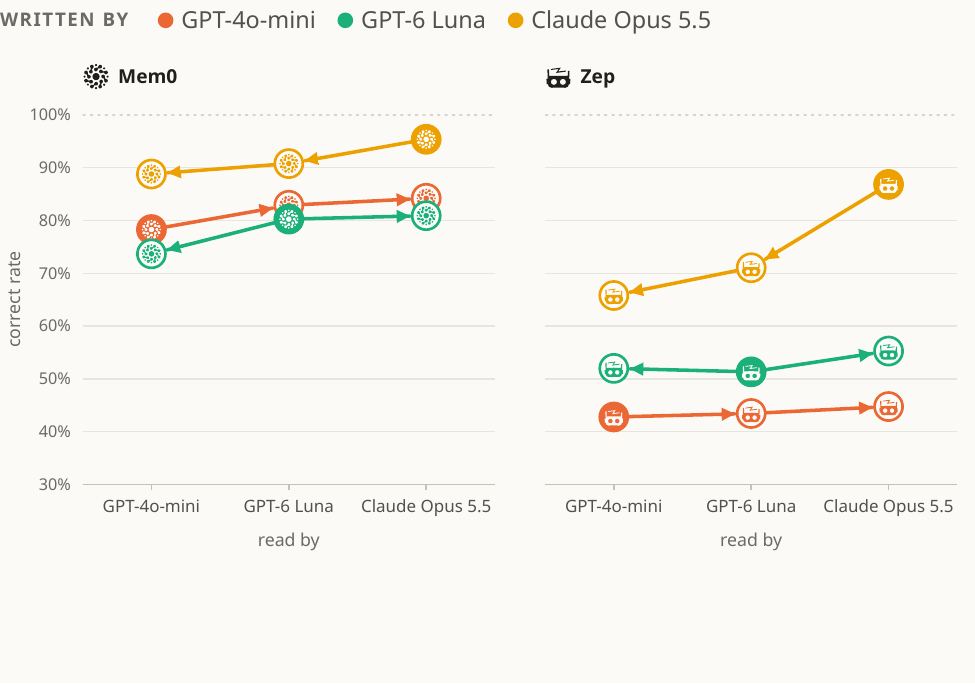}\par\vspace{2pt}\figrule
\caption{Writer and reader. Each colour is one store, written by that model. The filled badge is the writer reading its own store, and arrows run from it to the other two readers.}
\label{fig:cross}
\end{figure}

The model that writes the store limits every reader. Mem0 and Zep write their store with the same model that later reads it. Retrieval calls no language model, so the reader can be swapped: each model answers from the retrievals of the others' stores (Figure~\ref{fig:cross}, Table~\ref{tab:cross}).

A stronger reader does not recover what the writer left out. On Zep's GPT-4o-mini graph, GPT-4o-mini, GPT-6 Luna and Claude Opus 5.5 score 43\%, 43\% and 45\%; on the graph Claude Opus 5.5 wrote, 66\%, 71\% and 87\%. Mem0 keeps the order with smaller gaps: 78\%, 83\% and 84\% from GPT-4o-mini's store, 89\%, 91\% and 95\% from Claude Opus 5.5's. This agrees with work where verbatim chunks beat extracted facts on long conversations~\citep{an2026fidelity}.

\section{With Claude Opus 5.5, Pi tops the frontier}
\label{sec:pi}

Pi with Claude Opus 5.5 is the top of the frontier in Figure~\ref{fig:frontier}. No other cell scores higher: Pi scores 99\% for \$6.95 a rollout, while full context with the same model scores 95\% for \$17.84. Pi reads about 10,000 input tokens a question, 7,900 of them cached, against 28,900 for full context. Codex and Claude Code score 98\% and 97\% for \$25.41 and \$28.41; HarnessTax explains why the same model costs so differently across harnesses~\citep{arena2026harnesstax}.

Its one miss is LoCoMo's mistake. The question asks what kind of painting Caroline shared with Melanie on 13 October 2023, and the answer key says an abstract painting with blue streaks. In the conversation it is Melanie who shares that painting; Caroline shares a drawing. Pi says so and is marked wrong: ``Caroline shared a drawing, not a painting. \ldots{} The paintings shared that day (a sunset and an abstract piece) were Melanie's.'' With Claude Opus 5.5, full context, Codex and Claude Code point out the same mistake and are marked wrong too. On the other 151 questions, Pi with Claude Opus 5.5 is always right.

Photo captions account for most of Pi's lead over full context. With Claude Opus 5.5, Pi answers seven questions that full context misses and misses none that it answers. Six of the seven have their evidence in a turn that shares a photo, whose caption Pi could read and full context could not (Section~\ref{sec:setup}); so do all six such questions for Codex and all four for Claude Code. On the 88 questions without photo evidence, Pi scores 100\%, and full context, Mem0, Codex and Claude Code 99\% each (Table~\ref{tab:photos}). With the same input, then, Pi matches full context for less than half its cost. Whether searching the record answers more than reading all of it, this study does not settle.

\section{Discussion and limitations}
\label{sec:discussion}

Our results suggest that agent memory is a costly patch for a weak model and caps what any reader can answer. On this LoCoMo conversation, Mem0 helps GPT-4o-mini but not GPT-6 Luna or Claude Opus 5.5, a stronger reader does not recover what the writer left out, and with Claude Opus 5.5 Pi answers every sound question for less than full context costs.

The findings rest on a small study, and these limits apply:
\begin{itemize}
\item \textbf{Scale.} One conversation of LoCoMo's ten, and one run per cell. Differences of a few questions may not hold in a second run.
\item \textbf{Grading.} A GPT-4o-mini judge with a lenient rubric, and at least one wrong answer key.
\item \textbf{Photo captions.} The agents could read photo captions that the memory methods were not given (Section~\ref{sec:pi}). This touches comparisons between agents and memory methods, not those among the memory methods.
\item \textbf{Context length.} The history fits in every model's context window, so full context is always an option. Results may differ on other benchmarks such as LongMemEval~\citep{wu2024longmemeval} and on histories that do not fit.
\item \textbf{Settings and prices.} Mem0 and Zep run with their default extraction and retrieval, and other settings may do better. Costs are list prices and depend on each provider's caching.
\end{itemize}

Memory systems and harnesses can also be searched for automatically~\citep{zhang2025memevolve,lee2026metaharness}, but what is found is still fixed before the question arrives. More broadly, a strong model may need less designed memory and better tools: exact, semantic and vector search over the raw record, driven by the model itself~\citep{pi2026,codex2026}. Progress may then look less like another memory system and more like ripgrep after grep~\citep{gallant2016ripgrep}, or zvec-grep after ripgrep~\citep{zvec2026zg}. Designed memory may survive as a cost trade-off when many questions share one history (Figure~\ref{fig:cumulative}).

\bibliographystyle{unsrtnat}
\bibliography{refs}

\appendix

\section{Full results}
\label{app:results}

\begin{figure}[p]
\centering
\figrule\par\vspace{2pt}
\includegraphics[width=0.68\linewidth]{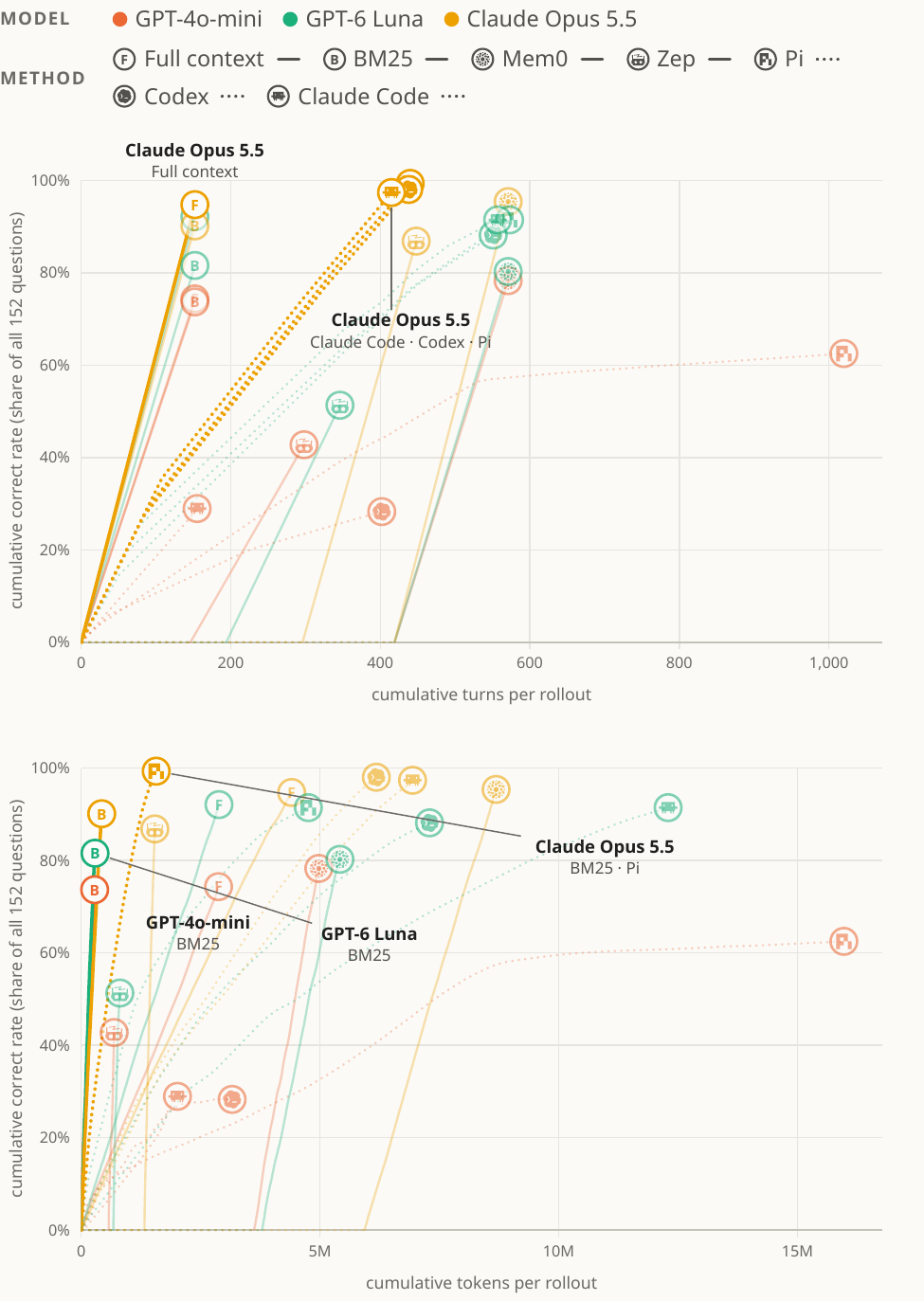}\par\vspace{2pt}\figrule
\caption{The curves of Figure~\ref{fig:cumulative} against turns (LLM calls, top) and tokens (bottom), on linear axes, the questions ordered by that amount. Strong lines are cells that no other cell beats on both that amount and correct answers.}
\label{fig:usage}
\end{figure}

Table~\ref{tab:results} gives every cell's cost and correct rate, Table~\ref{tab:cross} the writer--reader grid behind Figure~\ref{fig:cross}, and Table~\ref{tab:photos} the split by photo evidence. Figure~\ref{fig:usage} redraws Figure~\ref{fig:cumulative} against turns and tokens.

\begin{table}[p]
\centering
\caption{Per method and model. Cost is the list price of every model call except the judge: Build is paid once, before any question; Per q.\ is everything one question costs, retrieval included, averaged over the 152 questions. $\circ$: nothing is built. Effort: the model's reasoning effort; GPT-4o-mini has none. Correct: share of answers judged correct, in \%, per LoCoMo category (single-hop, multi-hop, temporal, open-domain); the first row gives each category's size.}
\label{tab:results}
\small\setlength{\tabcolsep}{5pt}%
\begin{tabular}{lllrrrrrrr}
\toprule
 & & & \multicolumn{2}{c}{Cost (\$)} & \multicolumn{5}{c}{Correct (\%)} \\
\cmidrule(lr){4-5}\cmidrule(lr){6-10}
Method & Model & Effort & Build & Per q. & Single & Multi & Temp. & Open & All \\
\midrule
\textit{Questions} & & & & & 70 & 32 & 37 & 13 & 152 \\
\midrule
Full context & GPT-4o-mini & \textendash &  $\circ$ & 0.0028 & 89 & 75 & 51 & 62 & 74 \\
 & GPT-6 Luna & high & $\circ$ & 0.0019 & 90 & 97 & 95 & 85 & 92 \\
 & Claude Opus 5.5 & high & $\circ$ & 0.117 & 93 & 94 & 97 & 100 & 95 \\
\midrule
BM25 & GPT-4o-mini & \textendash &  $\circ$ & 0.00028 & 86 & 50 & 73 & 69 & 74 \\
 & GPT-6 Luna & high & $\circ$ & 0.00021 & 89 & 63 & 89 & 69 & 82 \\
 & Claude Opus 5.5 & high & $\circ$ & 0.013 & 93 & 81 & 92 & 92 & 90 \\
\midrule
Mem0 & GPT-4o-mini & \textendash &  0.559 & 0.0013 & 87 & 75 & 65 & 77 & 78 \\
 & GPT-6 Luna & high & 0.449 & 0.0011 & 89 & 88 & 57 & 85 & 80 \\
 & Claude Opus 5.5 & high & 25.876 & 0.074 & 93 & 97 & 97 & 100 & 95 \\
\midrule
Zep & GPT-4o-mini & \textendash &  0.092 & 0.00012 & 51 & 34 & 30 & 54 & 43 \\
 & GPT-6 Luna & high & 0.103 & 0.00012 & 54 & 50 & 49 & 46 & 51 \\
 & Claude Opus 5.5 & high & 6.224 & 0.0077 & 83 & 94 & 89 & 85 & 87 \\
\midrule
Pi & GPT-4o-mini & \textendash &  $\circ$ & 0.016 & 74 & 59 & 35 & 85 & 63 \\
 & GPT-6 Luna & high & $\circ$ & 0.0032 & 96 & 81 & 92 & 92 & 91 \\
 & Claude Opus 5.5 & high & $\circ$ & 0.046 & 99 & 100 & 100 & 100 & 99 \\
\midrule
Codex & GPT-4o-mini & \textendash &  $\circ$ & 0.0031 & 30 & 22 & 24 & 46 & 28 \\
 & GPT-6 Luna & high & $\circ$ & 0.0050 & 90 & 84 & 89 & 85 & 88 \\
 & Claude Opus 5.5 & high & $\circ$ & 0.167 & 99 & 100 & 95 & 100 & 98 \\
\midrule
Claude Code & GPT-4o-mini & \textendash &  $\circ$ & 0.0020 & 34 & 22 & 8 & 77 & 29 \\
 & GPT-6 Luna & high & $\circ$ & 0.0082 & 96 & 78 & 95 & 92 & 91 \\
 & Claude Opus 5.5 & high & $\circ$ & 0.187 & 99 & 94 & 97 & 100 & 97 \\
\bottomrule
\end{tabular}

\end{table}

\begin{table}[p]
\centering
\caption{Writer and reader. Correct rate (\%), the reader answering from the writer's store and its retrievals. Bold: one model writes and reads (Table~\ref{tab:results}).}
\label{tab:cross}
\small
\small\setlength{\tabcolsep}{7pt}%
\begin{tabular}{lrrrrrr}
\toprule
 & \multicolumn{3}{c}{Mem0, read by} & \multicolumn{3}{c}{Zep, read by} \\
\cmidrule(lr){2-4}\cmidrule(lr){5-7}
Written by & GPT-4o-mini & GPT-6 Luna & Opus 5.5 & GPT-4o-mini & GPT-6 Luna & Opus 5.5 \\
\midrule
GPT-4o-mini & \textbf{78} & 83 & 84 & \textbf{43} & 43 & 45 \\
GPT-6 Luna & 74 & \textbf{80} & 81 & 52 & \textbf{51} & 55 \\
Claude Opus 5.5 & 89 & 91 & \textbf{95} & 66 & 71 & \textbf{87} \\
\bottomrule
\end{tabular}

\end{table}

\begin{table}[p]
\centering
\caption{Correct rate (\%) on the 64 questions with evidence in a turn that shares a photo, whose caption only the agents could read, and on the other 88.}
\label{tab:photos}
\small\setlength{\tabcolsep}{6pt}%
\begin{tabular}{lrrrrrr}
\toprule
 & \multicolumn{2}{c}{GPT-4o-mini} & \multicolumn{2}{c}{GPT-6 Luna} & \multicolumn{2}{c}{Claude Opus 5.5} \\
\cmidrule(lr){2-3}\cmidrule(lr){4-5}\cmidrule(lr){6-7}
Method & Photo (64) & Other (88) & Photo (64) & Other (88) & Photo (64) & Other (88) \\
\midrule
Full context & 66 & 81 & 91 & 93 & 89 & 99 \\
BM25 & 61 & 83 & 78 & 84 & 84 & 94 \\
Mem0 & 70 & 84 & 72 & 86 & 91 & 99 \\
Zep & 33 & 50 & 53 & 50 & 77 & 94 \\
Pi & 53 & 69 & 88 & 94 & 98 & 100 \\
Codex & 22 & 33 & 84 & 91 & 97 & 99 \\
Claude Code & 17 & 38 & 91 & 92 & 95 & 99 \\
\bottomrule
\end{tabular}

\end{table}

\clearpage
\section{Protocols}
\label{app:protocols}

Mem0 OSS 2.1.0 writes one chronological turn at a time and retrieves $k=200$ without reranking; dates are prepended to the text. Zep runs Graphiti OSS 0.30.2 on Kuzu, one session per episode, with hybrid edge search (top 20). The memory methods' GPT-6 Luna and Claude Opus 5.5 requests set reasoning effort high.

Each agent question is a Harbor 0.24.0 task~\citep{harbor2026}: a fresh container holding only the conversation, exactly as in the official file, without questions or answers. The agent's last message is its answer. While the agent runs, the container can reach only the model endpoint. GPT-6 Luna and Claude Opus 5.5 run at reasoning effort high, through a proxy that translates the API where an agent does not speak the model's own; GPT-4o-mini has no effort setting. Codex and Claude Code have web search off; Pi has none. A question may take 20 minutes, and Pi and Claude Code at most 100 turns. With GPT-6 Luna, Codex picks its code mode: one JavaScript tool that calls the shell. A trial that fails or ends without an answer is run again, and only the accepted trial is counted.

\section{Prompts}
\label{app:prompts}

Each prompt is copied from the code the runs called; fields in braces are filled per question.

\medskip\noindent\textbf{Answer, system message.}
\begin{lstlisting}
You are a helpful assistant that can answer questions based on the provided context.If the question involves timing, use the conversation date for reference.Provide the shortest possible answer.Use words directly from the conversation when possible.Avoid using subjects in your answer.
\end{lstlisting}

\medskip\noindent\textbf{Answer, user message.}
\begin{lstlisting}
# Question:
{{QUESTION}}

# Context:
{{CONTEXT}}

# Short answer:
\end{lstlisting}

\medskip\noindent\textbf{Pi, Codex and Claude Code, user message.}
\begin{lstlisting}
Answer a question about the conversation in `conv-26/conv.json`. If the question involves timing, use the conversation date for reference. Provide the shortest possible answer. Use words directly from the conversation when possible. Avoid using subjects in your answer. Your final message is taken as the answer.

Question: {question}
\end{lstlisting}

\medskip\noindent\textbf{Judge.}
\begin{lstlisting}
Your task is to label an answer to a question as ’CORRECT’ or ’WRONG’. You will be given the following data:
    (1) a question (posed by one user to another user),
    (2) a ’gold’ (ground truth) answer,
    (3) a generated answer
which you will score as CORRECT/WRONG.

The point of the question is to ask about something one user should know about the other user based on their prior conversations.
The gold answer will usually be a concise and short answer that includes the referenced topic, for example:
Question: Do you remember what I got the last time I went to Hawaii?
Gold answer: A shell necklace
The generated answer might be much longer, but you should be generous with your grading - as long as it touches on the same topic as the gold answer, it should be counted as CORRECT.

For time related questions, the gold answer will be a specific date, month, year, etc. The generated answer might be much longer or use relative time references (like "last Tuesday" or "next month"), but you should be generous with your grading - as long as it refers to the same date or time period as the gold answer, it should be counted as CORRECT. Even if the format differs (e.g., "May 7th" vs "7 May"), consider it CORRECT if it's the same date.

Now it's time for the real question:
Question: {question}
Gold answer: {gold_answer}
Generated answer: {generated_answer}

First, provide a short (one sentence) explanation of your reasoning, then finish with CORRECT or WRONG.
Do NOT include both CORRECT and WRONG in your response, or it will break the evaluation script.

Just return the label CORRECT or WRONG in a json format with the key as "label".
\end{lstlisting}

\end{document}